\pdfoutput=1
\pdfoutput=1
\documentclass[11pt]{article}

\usepackage[]{emnlp2021}

\usepackage{times}
\usepackage{latexsym}

\usepackage[T1]{fontenc}

\usepackage[utf8]{inputenc}

\usepackage{amsmath}
\usepackage{amssymb}
\usepackage{graphicx}
\usepackage{subfigure}
\usepackage{multirow}
\usepackage{booktabs}

\usepackage{microtype}

\title{Rumor Detection on Twitter with Claim-Guided Hierarchical Graph Attention Networks}

\author{Hongzhan Lin$^{1,2}$, Jing Ma$^{2,*}$, Mingfei Cheng$^1$, Zhiwei Yang$^3$, Liangliang Chen$^1$, Guang Chen$^{1,}$\thanks{{\; Corresponding authors.}} \\
        $^1$Beijing University of Posts and Telecommunications, Beijing, China \\ $^2$Hong Kong Baptist University, Hong Kong SAR, China \\ $^3$Jilin University, Changchun, China \\
        \texttt{\{linhongzhan,mingfeicheng,outside,chenguang\}@bupt.edu.cn}\\
        \texttt{majing@comp.hkbu.edu.hk, yangzw18@mails.jlu.edu.cn}}

\begin{document}
\maketitle
\begin{abstract}
Rumors are rampant in the era of social media. Conversation structures provide valuable clues to differentiate between real and fake claims. However, existing rumor detection methods are either limited to the strict relation of user responses or oversimplify the conversation structure. In this study, to substantially reinforces the interaction of user opinions while alleviating the negative impact imposed by irrelevant posts, we first represent the conversation thread as an undirected interaction graph. We then present a Claim-guided Hierarchical Graph Attention Network for rumor classification, which enhances the representation learning for responsive posts considering the entire social contexts and attends over the posts that can semantically infer the target claim. Extensive experiments on three Twitter datasets demonstrate that our rumor detection method achieves much better performance than state-of-the-art methods and exhibits a superior capacity for detecting rumors at early stages.

\end{abstract}

\section{Introduction}

Rumor is one type of social diseases in the era of social media. 
The spread of false rumors has a far-reaching destructive impact on both society and individuals~\cite{ma2019detect}. For instance, the global COVID-19 pandemic has created fertile soil for the widespread of various rumors, conspiracy theories, hoaxes, and fake news, heavily 
disrupting people's peaceful lives and leading to unprecedented information chaos. A strange, new rumor claiming that ``wearing a mask to prevent the spread of COVID-19 is unnecessary because the disease can also be spread via farts"
\footnote{\url{https://www.snopes.com/fact-check/farting-negate-covid19-masks/}} 
may mislead masses to belittle the importance of those potentially life-saving masks in epidemic prevention. Therefore, it is necessary to develop automatic approaches to facilitate rumor detection, especially amid crises.

\begin{figure*}[ht]
\centering
\subfigure[Sample of a conversation thread from a false rumor]{
\begin{minipage}[t]{0.6\linewidth}
\centering
\scalebox{1.63}{\includegraphics[width=6cm]{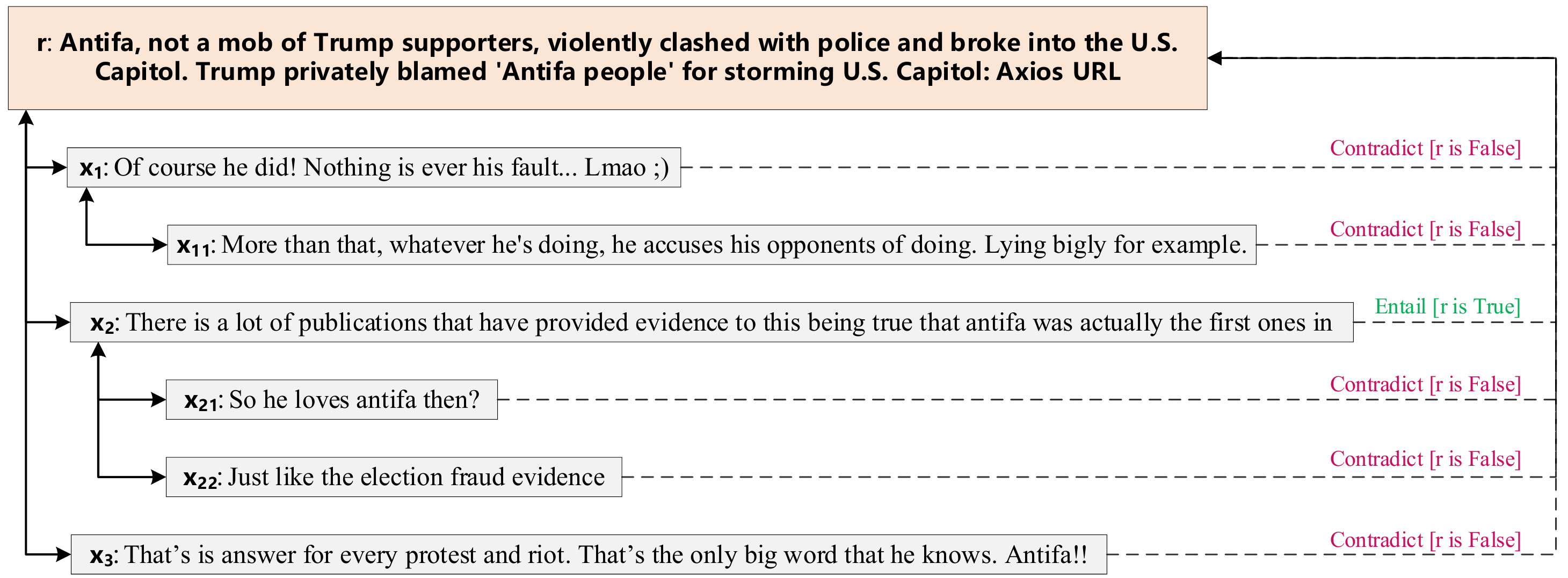}}
\label{fig:conversation}
\end{minipage}%
}%
\subfigure[Undirected interaction graph]{
\begin{minipage}[t]{0.4\linewidth}
\centering
\scalebox{0.90}{\includegraphics[width=6cm]{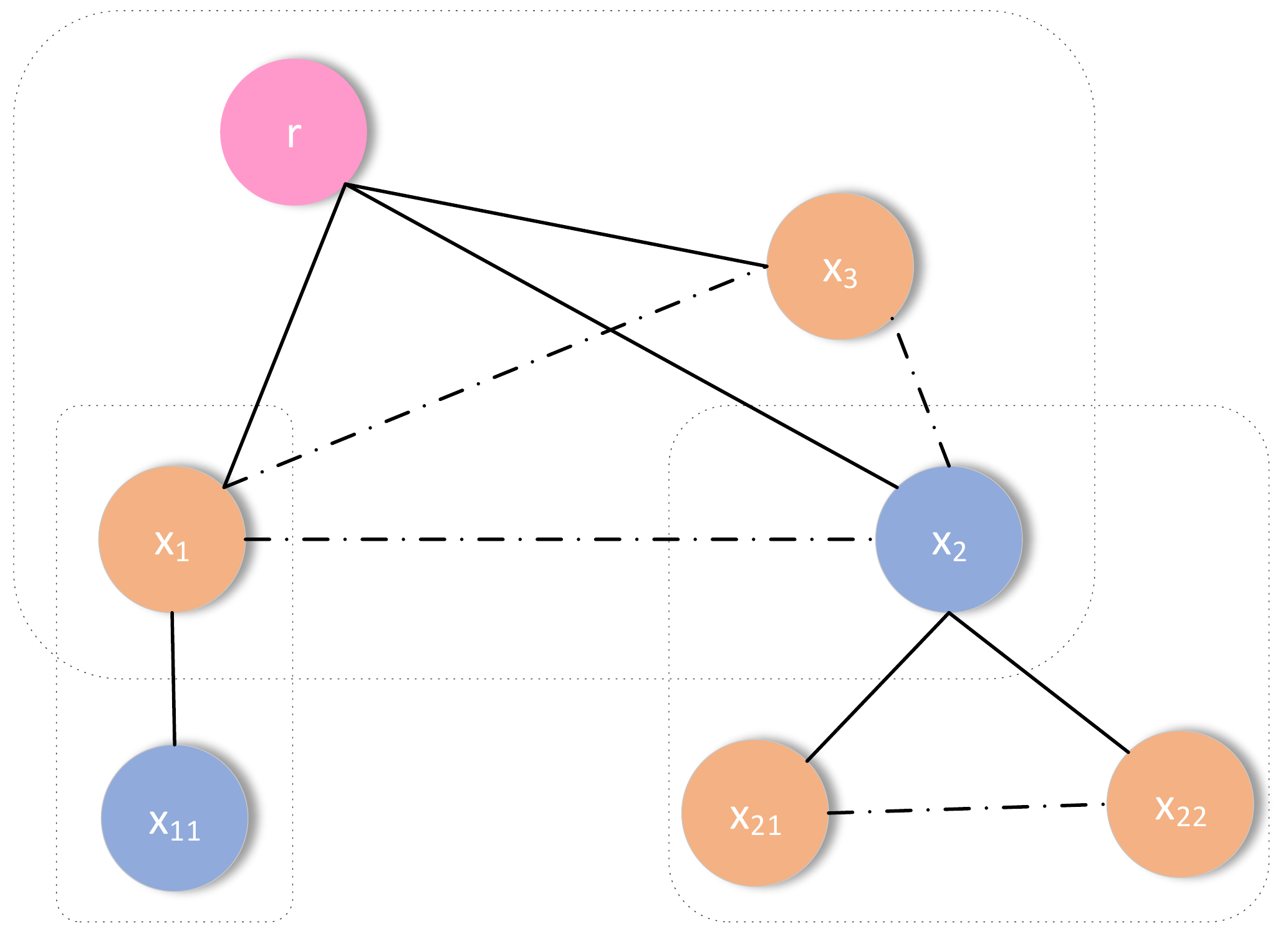}}
\label{fig:interaction}
\end{minipage}%
}%
\centering
\caption{(a) A motivating example: A false rumor widely spread on Twitter. (b) The undirected interaction graph for modeling the conversation thread. Blue nodes support or confirm the replied node, while orange nodes refute. For clarity's sake, we distinguish the responsive/sibling relationships between nodes with solid/chain lines.}
\label{fig:1}
\vspace{-0.5cm}
\end{figure*}

Social psychology literature defines a rumor as a story or a statement whose truth value is unverified or intentionally false~\cite{difonzo2007rumor}. Rumor detection aims to determine the veracity of a given story or statement. For automating rumor detection, previous studies focus on text mining from sequential microblog streams with supervised classifiers based on feature engineering~\cite{castillo2011information,yang2012automatic,kwon2013prominent,liu2015real, ma2015detect} and feature learning~\cite{ma2016detecting, yu2017convolutional}. 
The interactions among users generally show conductive to provide useful clues for debunking rumors. Structured information is generally observed on social media platforms such as Twitter. 
Structure-based methods~\cite{ma2017detect,ma2018rumor} are thus proposed to capture the interactive characteristics of rumor diffusion. We discuss briefly two types of state-of-the-art approaches: Transformer-based~\cite{khoo2020interpretable,ma2020debunking} and Directed GCN-based~\cite{bian2020rumor} models. 

\citet{khoo2020interpretable} exploited post-level self-attention networks to model long-distance interactions between any pair of tweets even irrelevant. \citet{ma2020debunking} further presented a tree-transformer model to make pairwise comparisons among the posts in the same subtree hierarchically, which better utilizes tree-structured user interactions in the conversation thread. \citet{bian2020rumor} utilized graph convolutional networks (GCNs) to encode directed conversation trees hierarchically. The structure-based methods however represent the conversation as a directed tree structure, following the bottom-up or top-down information flows. But such kind of structure, considering directed responsive relation, cannot enhance the representation learning of each tweet by aggregating information in parallel from the other informative tweets.

In this paper, we firstly represent the conversation thread as an undirected interaction graph, which allows full-duplex interactions between posts with responsive parent-child or sibling relationships so that the rumor indicative features from neighbors can be fully aggregated and the interaction of user opinions can be reinforced. Intuitively, we exemplify a false rumor claim and illustrate its propagation on Twitter in Figure~\ref{fig:conversation}. We observe that a group of tweets is triggered to reply to the same post (i.e., parent post) in the conversation thread. 
As users share opinions, conjectures, and evidence, inaccurate information on social media can be ``self-checked" by making a comparison with correlative tweets \cite{zubiaga2018detection}. 
In order to lower the weight of inaccurate responsive information (e.g., the supportive post $x_2$ toward the false claim $r$), coherent opinions need to be captured by comparing all responsive posts toward the same post. To achieve this, our proposed interaction topology as shown in Figure~\ref{fig:interaction} takes the correlations between sibling nodes such as the dotted box portion into account. On the other hand, by leveraging the intrinsic structural property of graph-based modeling, the undirected graph allows each tweet to learn the representation by aggregating features from all its informative neighbors. In this way, information association between nodes in the conversation can be adaptively propagated to each other along the responsive parent-child or sibling relationships while avoiding the negative impact of irrelevant interactions such as the comparison between $x_{11}$ and $x_{21}$ in Figure~\ref{fig:conversation}.

Moreover, previous studies show that it is critical to strengthen the semantic inference capacity between posts and the claim based on textual entailment reasoning~\cite{ma2019sentence}, so that we could semantically infer the claim by implicitly excavating textual inference relations such as entail, contradict, and neutral. We hypothesis that all the informative posts should be developed and extended around the content of the claim, i.e., the potential and implicit target to be checked. Therefore, the claim content is significant to catch informative tweets, such as that 
in Figure~\ref{fig:conversation}, it is observed that $x_{22}$ satirizes the opinion expressed in $x_2$, but its contextual information is limited. Integrating claim information for claim-aware representations could not only enrich the semantic context of $x_{22}$, but also enable it to better guard the consistency of topics when interacting with other nodes such as $x_2$ and $x_{21}$. 


To this end, we propose a novel Claim-guided Hierarchical Graph Attention Network (ClaHi-GAT) for detecting rumors on Twitter, which not only enhances the representation learning for posts by taking the entire conversation context but also attends over the subset of informative posts.
More specifically, we firstly model the conversation thread of a claim as an undirected interaction graph. To flexibly deal with the interaction of node information and the association of the global structure of the graph, we propose ClaHi-GAT to embed the undirected interaction graph. Different from standard graph attention networks (GATs)~\cite{velivckovic2017graph}, we design a claim-guided hierarchical attention mechanism at both post and event level to attend over informative posts by considering the coherent attitude and semantic inference strength toward the claim. As a result, the post-level representation is enhanced by the claim-aware attention weights obtained based on the textual content of the claim. 
Finally, we utilize an inference-based attention layer to implicitly capture the inference relation between the claim and the selected informative posts for rumor prediction at the event-level. 
We conduct extensive experiments on THREE public Twitter datasets and demonstrate that our proposed ClaHi-GAT model yields outstanding improvements over the state-of-the-art baselines with a large margin, and our method performs particularly well on early rumor detection which is crucial for timely intervention and debunking. The main contributions of this paper are three-fold:
\vspace{-0.21cm}
\begin{itemize}
\item To our best knowledge, this is the first study of representing conversation structure as an undirected interaction graph. The graph attention-based representation achieves significant improvements over state-of-the-art methods that rely on bottom-up/top-down tree structure. 
\vspace{-0.21cm}
\item We propose a novel ClaHi-GAT model to represent both tweet contents and the interaction graph into a latent space, which captures multi-level rumor indicative features via a claim-aware attention at the post level and an inference-based attention at the event level.
\vspace{-0.21cm}
\item Experimental results show that our model achieves superior performance on three real-world Twitter benchmarks for both rumor classification and early detection tasks.
\end{itemize}

\section{Related Work}
Pioneer studies for automatic rumor detection focus on features crafted from post contents, user profiles, and propagation patterns to learn a supervised classifier \cite{castillo2011information, yang2012automatic, liu2015real}. Subsequent studies were then conducted to engineer new features such as those representing rumor diffusion and cascades \cite{kwon2013prominent, friggeri2014rumor, hannak2014get}. \citet{ma2015detect} extended their model with a large set of chronological social context features. These approaches typically require heavy preprocessing and feature engineering. 

\citet{zhao2015enquiring} relieved the engineering effort by using a set of regular expressions (such as “really?”, “not true”, etc) to find questing and denying tweets, but the oversimplified approach suffered from very low recall. \citet{ma2016detecting} and \citet{yu2017convolutional} respectively utilized recurrent neural networks (RNNs) and convolutional neural networks (CNNs) to learn the representations from tweets content based on time series. 
\citet{guo2018rumor} proposed a hierarchical attention model that captures important clues from the social context of a rumorous event at the post and sub-event levels. \citet{jin2016news} exploited the conflicting viewpoints in a credibility propagation network for verifying news stories propagated among the tweets. However, these approaches cannot embed features reflecting how posts are propagated and require careful data segmentation to prepare for time sequences.

\begin{figure*}[t]
\centering
\scalebox{0.8}{\includegraphics[width=20cm]{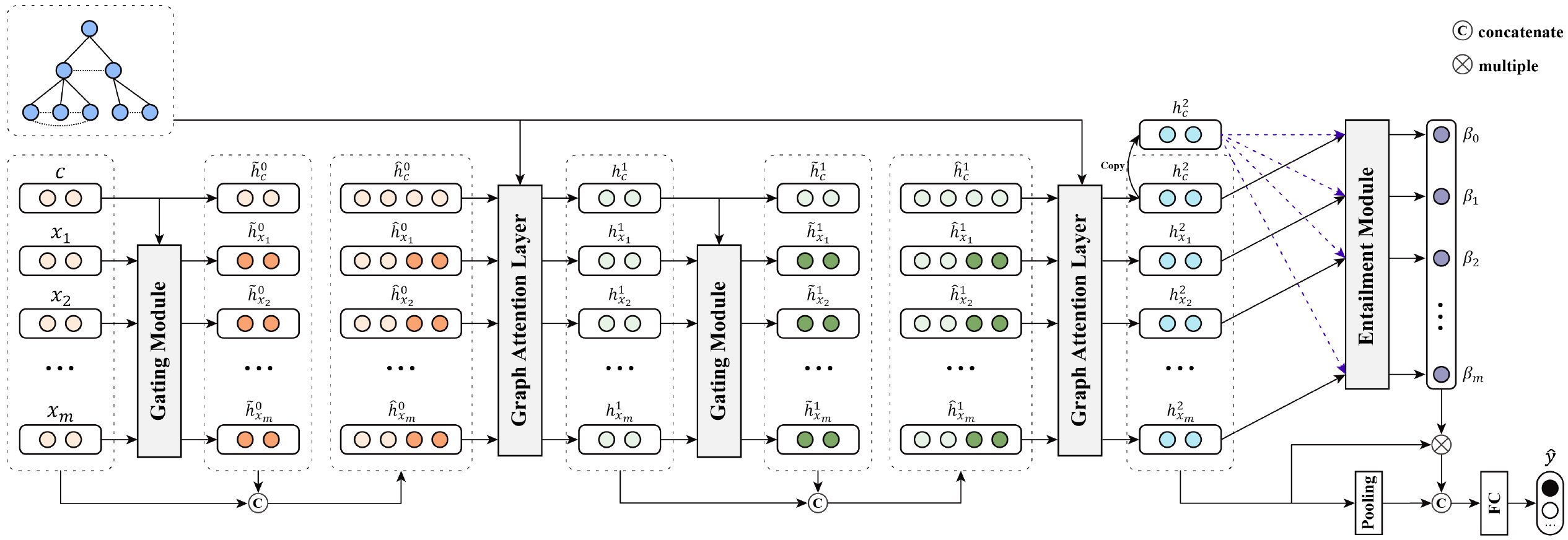}}
\caption{The architecture of our proposed Claim-guided Hierarchical Graph Attention Networks.}
\label{fig:method}
\vspace{-0.5cm}
\end{figure*}

To extract useful clues jointly from content semantics and propagation structures, 
\citet{wu2015false} proposed a hybrid SVM classifier to capture both flat and propagation patterns for detecting rumors on Sina Weibo. \citet{ma2017detect} used Tree Kernel to capture the similarity of propagation trees in order to identify different types of rumors on Twitter. \citet{ma2018rumor} presented tree-structured recursive neural networks (RvNN) to jointly generate the representation of a propagation tree based on the post contents and their propagation structure. 
More recently, 
\citet{khoo2020interpretable} proposed to model potential dependencies between any two microblog posts with the post-level self-attention networks, which is too vulnerable to avoid the negative impact of interactions among irrelevant posts. \citet{ma2020debunking} treated transformer as the unit of the tree structure to further enhance the representation learning but its running time is sensitive to conversation's depth. \citet{bian2020rumor} used GCNs~\cite{kipf2016semi} to encode the bi-directional conversation trees for higher-level representations.

In recent years, GATs have demonstrated superior performance in a variety of NLP tasks, such as text classification~\cite{linmei2019heterogeneous}, machine reading~\cite{zheng2020document}, recommendation system~\cite{wang2019kgat}, modeling knowledge graph~\cite{cui2020deterrent} and social network bias~
\cite{yuan2019jointly, huang2020heterogeneous}, etc. Different from these previous works, in this paper, we attempt to learn graph attention-based embeddings that attend to user interactions from community response for rumor detection. 

\section{Problem Statement}
We define a Twitter rumor detection dataset as a set of events $\mathbb{C} = \{C_1, C_2, ... , C_{|\mathbb{C}|}\}$, where each event $C_\tau$ corresponds to a claim $\textbf{c}$, composed of ideally all its relevant responsive tweets in chronological order, i.e., $C_\tau = \{\textbf{c}, \textbf{x}_{1},\textbf{x}_{2},...,\textbf{x}_{m} \}$, where $\textbf{c}$ can also be denoted as $\textbf{x}_{0}$ and $m$ is the number of responsive tweets in the conversation thread. Note that although the tweets are notated sequentially, there are connections among them based on their reply or repost relationships. So most previous works represent the conversation thread as a directed tree structure~\cite{wu2015false,ma2017detect,ma2018rumor,khoo2020interpretable}.



We formulate the task of rumor detection as a supervised classification problem that learns a classifier $f$ from the labeled claims, that is, $f: C_\tau \rightarrow Y_\tau$, where $Y_\tau$ takes one of the classes defined by the specific dataset:
\vspace{-0.21cm}
\begin{itemize}
\item Binary labels: rumor and non-rumor, which simply predicts a claim as rumor or not;
\vspace{-0.21cm}
\item Finer-grained labels: non-rumor, false rumor, true rumor, and unverified rumor, which makes rumor detection a more challenging classification problem \cite{ma2017detect, 2016Analysing}.
\end{itemize}

\noindent{\textbf{Undirected Interaction Graphs Construction.}} On Twitter, each set of responsive posts triggered by the same post contains distinct rumor-indicative patterns \cite{ma2017detect}. It is worth noting that we consider interactions not just between responsive parent-child nodes, but also those with the sibling relationship, for better feature aggregation from the informative tweets.
To explore the full-duplex interaction patterns between responsive parent-child nodes or sibling nodes, we model the interaction topology among tweets as an undirected graph $\mathcal{G}=\langle \mathcal{V}, {\mathcal{E}}\rangle$ 
for an undetermined event $C_\tau$, as exemplified in Figure~\ref{fig:interaction}, where $\mathcal{V} = C_{\tau}$ that consists of all relevant posts as nodes and ${\mathcal{E}}$ refers to a set of undirected edges corresponding to the interactions between the nodes in $\mathcal{V}$. For example, for any $\textbf{x}_i,\textbf{x}_j \in \mathcal{V}$, $\textbf{x}_i \to \textbf{x}_j$ and $\textbf{x}_j \to \textbf{x}_i$ exist if they have responsive parent-child or sibling relationships. 

\section{Claim-guided Hierarchical Graph Attention Networks}
In this section, we introduce our Claim-guided Hierarchical Graph Attention Networks to embed the undirected interaction graph for rumor detection. The proposed neural network consists of two attention mechanisms, i.e., a Graph Attention to capture the importance of different neighboring tweets, and a claim-guided hierarchical attention to enhance post content understanding. Figure~\ref{fig:method} illustrates an overview of our proposed model, which will be depicted in the following subsections.


\subsection{Graph Attention Networks}
\label{GATs}
The core idea of GATs is to enhance the representation of responsive posts, which assign various levels of importance to neighboring posts, rather than treating all of them with equal importance, as is done in the GCN model. 
Our intuition for applying GATs to embed undirected interaction graphs is to reduce the weights of noisy information. 

Given a tweet $\textbf{x}_i$, we utilize a bi-directional LSTM encoder over its involved word sequence which is represented by pre-trained word embeddings. We then obtain the post-level representation using the last hidden state of the bi-directional LSTM. We thus denote the event as a matrix, i.e., $X = {[c, {x}_1, {x}_2, \cdots, {x}_{|\mathcal{V}|-1}]}^{\top}$, where $c,{x}_i \in \mathbb{R}^d$ respectively denotes the $d$-dimensional embedding of the claim and each responsive tweet. 

In order to encode structural contexts to improve the post-level representation by adaptively aggregating more informative signals from neighboring tweets, we utilize self-attention to model the interactions between one tweet and its neighboring tweets in $\mathcal{G}$. So the attention coefficients would correlate to the impact of neighbors on the current tweet. Specifically, the input for the calculation is a set of vectors, ${H^{(l)}}= [{h}_{c}^{(l)}, {h}_{x_1}^{(l)}, {h}_{x_2}^{(l)},...,{h}_{x_{|\mathcal{V}|-1}}^{(l)}]^{\top}$ that denotes the hidden representations of nodes at the $l$-th layer and ${h}_{c}^{(l)}$ can also be denoted as ${h}_{x_0}^{(l)}$. Initially, $H^{(0)} = X$. The attention coefficient can be computed as follows:
{\setlength{\abovedisplayskip}{0.1cm}
\setlength{\belowdisplayskip}{0.1cm}
\begin{equation}
    \begin{aligned}
    \alpha_{i,j}^{(l)}&=\emph{Atten}(h_{x_i}^{(l)},h_{x_j}^{(l)})\\&=\frac{exp{(\phi(a^\top[W^{(l)}{h}_{x_i}^{(l)}||W^{(l)}{h}_{x_j}^{(l)}]))}}{\sum\limits_{j\in \mathcal{N}_i}exp{(\phi(a^\top[W^{(l)}{h}_{x_i}^{(l)}||W^{(l)}{h}_{x_{j}}^{(l)}]))}}
    \end{aligned}
    \label{equ1}
\end{equation}} where $\alpha_{i,j}^{(l)}$ indicates the importance of tweet $\textbf{x}_j$ to $\textbf{x}_i$, $a$ is a weight vector, $W^{(l)}$ is a layer-specific trainable transformation matrix, $||$ means ``concatenate" operation, $\mathcal{N}_i$ contains $\textbf{x}_i$'s one-hop neighbors and $\textbf{x}_i$ itself, $\phi$ denotes the activation function, such as LeakyReLU\cite{2014Rich}. Then the layer-wise propagation rule is defined as:
{\setlength{\abovedisplayskip}{0.1cm}
\setlength{\belowdisplayskip}{0.1cm}
\begin{equation}
    h_{x_i}^{(l+1)}= ReLU(\sum\limits_{j\in \mathcal{N}_i}\alpha_{i,j}^{(l)}W^{(l)}h_{x_j}^{(l)})
    \label{equ2}
\end{equation}} 
After that, multi-head attention is introduced to expand the channel of self-attention and stabilize the learning process~\cite{vaswani2017attention}. Thus Eq.\ref{equ2} would be extended to the multi-head attention process of concatenating $K$ attention heads:
{\setlength{\abovedisplayskip}{0.1cm}
\setlength{\belowdisplayskip}{0.1cm}
\begin{equation}
    h_{x_i}^{(l+1)}= \mathop{\Vert}\limits_{k=1}^{K} ReLU(\sum\limits_{j\in \mathcal{N}_i}\alpha_{i,j}^{(l,k)}W_k^{(l)}h_{x_j}^{(l)})
    \label{equ3}
\end{equation}} 
where $h_{x_i}^{(l+1)}$ denotes the hidden representations of the tweet $\textbf{x}_i$ at the ($l$+1)-th layer. $\alpha_{i,j}^{(l,k)}$ is a normalized attention coefficient calculated by the $k$-th head at the $l$-th layer, and $W_k^{(l)}$ represents the corresponding linear transformation matrix. After going through an $L$-layer GAT, the output embedding in the final layer is calculated using averaging, instead of the concatenation operation:
{\setlength{\abovedisplayskip}{0.1cm}
\setlength{\belowdisplayskip}{0.1cm}
\begin{equation}
    h_{x_i}^{(L)}= ReLU(\frac{1}{K}\sum\limits_{k=1}^{K}\sum\limits_{j\in \mathcal{N}_i}\alpha_{i,j}^{(l',k)}W_k^{(l')}h_{x_j}^{(l')})
    \label{equ4}
\end{equation}} where $l'=L-1$, $h_{x_i}^{(L)}$ is the refined node representation of $\textbf{x}_i$ after aggregating information from the other informative tweets. 
Here we employ mean-pooling operators to jointly capture the opinions expressed in the whole conversation, which is obtained based on the refined node representation:
{\setlength{\abovedisplayskip}{0.1cm}
\setlength{\belowdisplayskip}{0.1cm} 
\begin{equation}
    \bar{s} = \emph{mean-pooling}(H^{(L)})
    \label{equ5}
\end{equation}} where $\bar{s}$ is the mean-pooled representation of the entire graph. 

\subsection{Claim-guided Hierarchical Attention}
On top of the GATs, we further propose the claim-guided hierarchical attention mechanism to strengthen the topical coherence and semantic inference for our model.

\noindent{\textbf{Post-level Attention.}} To make full use of abundant information in the claim and prevent off-topic coherence that deviates from the claim’s focus, we exploit a gating module to endow the model with the capacity of deciding how much information it should accept from the claim for better guiding the importance allocation of the related post in the neighborhood. The claim-aware representation could be obtained as follows:
{\setlength{\abovedisplayskip}{0.1cm}
\setlength{\belowdisplayskip}{0.1cm}
\begin{equation}
    \begin{aligned}
    g_{c\rightarrow x_i}^{(l)}&={sigmoid}(W_g^{(l)}{h}_{x_i}^{(l)}+U_g^{(l)}{h}_{c}^{(l)})\\
    \tilde{h}_{x_i}^{(l)} &= g_{c\rightarrow x_i}^{(l)}\odot {h}_{x_i}^{(l)} +(1-g_{c\rightarrow x_i}^{(l)})\odot {h}_{c}^{(l)}
    \end{aligned}
    \label{equ7}
\end{equation}} where $g_{c\rightarrow x_i}^{(l)}$ is the gate vector at the $l$-th layer, with trainable parameters $W_g^{(l)}$ and $U_g^{(l)}$. We omit the bias to avoid notation clutter. $\odot$ denotes Hadamard product. Then we concatenate the claim-aware representation with the original representation to feed into Eq.\ref{equ1} for a refined claim-aware attention weight:
{\setlength{\abovedisplayskip}{0.1cm}
\setlength{\belowdisplayskip}{0.1cm}
\begin{equation}
    \begin{aligned}
    \hat{h}_{x_i}^{(l)}&=[\tilde{h}_{x_i}^{(l)}||h_{x_i}^{(l)}]\\
    \hat{\alpha}_{i,j}^{(l)}&=\emph{Atten}(\hat{h}_{x_i}^{(l)},\hat{h}_{x_j}^{(l)})
    \end{aligned}
\end{equation}}

Note that in this way, we update the raw representation and attention score $h$, $\alpha$ fed into Eq.~\ref{equ2}-\ref{equ4} with the refined representation and attention score $\hat{h}$, $\hat{\alpha}$, so that our model can determine the verdict of a claim more reasonably with evidential posts taking the learned claim representation into account.

\noindent{\textbf{Event-level Attention.}} A natural argument against the prior GAT-mean-based model (see Section~\ref{GATs}) is that mean-pooling over the node vectors does not always make sense, since some nodes are more important than others for reasoning the veracity of the rumorous event. In order to strengthen the semantic inference capacity of our model, we propose an inference module at the event level to implicitly capture the entailment relations between the posts and the claim based on the Natural Language Inference (NLI) \cite{bowman2015large}.

Inspired by the matching scheme used in classical NLI models \cite{mou2015natural, yang2019simple}, given the output of the last graph attention layer, we conduct each such pair by integrating three matching functions between $h_c^{(L)}$ and $h_{x_i}^{(L)}$: 1) concatenation $[h_c^{(L)}||h_{x_i}^{(L)}]$; 2) element-wise product $h_{\emph{prod}}^{(L)}=h_c^{(L)}\odot h_{x_i}^{(L)}$; 3) absolute element-wise difference $h_{\emph{diff}}^{(L)}=|h_c^{(L)}-h_{x_i}^{(L)}|$. Afterwards, we can obtain a joint representation as:
{\setlength{\abovedisplayskip}{0.1cm}
\setlength{\belowdisplayskip}{0.1cm}
\begin{equation}
    h_{x_i}^{c}={tanh}\left(\emph{FC}([h_c^{(L)}|| h_{x_i}^{(L)}|| h_{\emph{prod}}^{(L)}||h_{\emph{diff}}^{(L)}])\right)
\end{equation}} We employ an attention over the output embeddings of the last graph attention layer to select inference-based informative posts, which is guided by the joint representation $h_{x_i}^{c}$. This yields:
{\setlength{\abovedisplayskip}{0.1cm}
\setlength{\belowdisplayskip}{0.1cm}
\begin{equation}
    \begin{aligned}
    b_i&={tanh}(\emph{FC}(h_{x_i}^{c}))\\
    \beta_i &=\frac{exp(b_i)}{\sum_{i}exp(b_i)}\\
    \hat{s}&=\sum\limits_{i}{\beta_i} h_{x_i}^{(L)}
    \end{aligned}
    \label{equ10}
\end{equation}} where $\beta_i$ is the normalized inference-based attention weight of $\textbf{x}_i$ for attaining the representation $\hat{s}$ of an entire graph. Lastly, we concatenate $\hat{s}$ with $\bar{s}$ and feed them into a fully-connected layer to get a low-dimensional veracity prediction vector:
{\setlength{\abovedisplayskip}{0.1cm}
\setlength{\belowdisplayskip}{0.1cm}
\begin{equation}
    \hat{y}={softmax}(\emph{FC}([\hat{s}||\bar{s}]))
\end{equation}}
where \emph{FC} means a fully-connected network.

\subsection{Model Training}
During model training, we exploit the cross-entropy loss of the predictions $\hat{y}$ and ground truth distributions $y$ over training data with the L2-norm. We set the number L of the graph attention layer as 2, and the head number K as 4. Parameters are updated through back-propagation \cite{collobert2011natural} with the Adam optimizer \cite{kingma2014adam}. The learning rate is initialized as 0.0005, and the dropout rate is 0.2. Early stopping \cite{yao2007early} is applied to avoid overfitting.

\section{Experiments}
\subsection{Datasets}
We conduct experiments on three public benchmarks, including Twitter15 \cite{ma2017detect}, Twitter16 \cite{ma2017detect}, and PHEME \cite{zubiaga2016learning}. The label of each event in Twitter15 and Twitter16 is annotated according to the veracity tag of the article in rumor debunking websites (e.g., snopes.com, Emergent.info, etc) \cite{ma2017detect}. Moreover, the fraction of different types of rumors is imbalanced in the real-world. For example, the number of real news usually far exceeds that of false rumors. Therefore, we resort to another public benchmark rumor dataset PHEME\footnote{\url{https://figshare.com/articles/dataset/PHEME_dataset_of_rumours_and_non-rumours/4010619}}, which is unbalanced and collected based on five real-world breaking news items. TWITTER (Twitter15\&16) datasets contain four labels: Non-rumor (NR), False Rumor (FR), True Rumor (TR), and Unverified Rumor (UR), while the unbalanced dataset PHEME collected based on five real-world breaking news items contains two binary labels: Rumor and Non-rumor. To evaluate the robustness of our model on complex responsive relations, we further split TWITTER datasets into TWITTER-S and TWITTER-D according to the conversation depth (TWITTER-S: $\leq 3$;  TWITTER-D: $\geq 4$) following \citet{ma2020debunking}. The full statistics of datasets and implementation details are shown in the appendix.

\subsection{Experimental Setup}


We compare our proposed model with the following baseline and state-of-the-art models: 1) \textbf{DTR}: A Decision-Tree-based Ranking model \cite{zhao2015enquiring} that identifies trending rumors by searching for inquiry phrases. 2) \textbf{DTC}: A decision tree-based model \cite{castillo2011information} that utilizes a combination of news characteristics. 3) \textbf{RFC}: A random forest classifier \cite{kwon2013prominent} with a set of hand-crafted features like linguistic and structure characteristics, etc. 4) \textbf{SVM-TK}: A SVM classifier that uses a Tree Kernel \cite{ma2017detect} which try to capture propagation structure via kernel learning. 5) \textbf{GRU-RNN}: A RNN-based model that learns temporal-linguistic patterns from user comments \cite{ma2016detecting}. 6) \textbf{RvNN}: A rumor detection approach based on tree-structured recursive neural networks \cite{ma2018rumor} with GRU units that learn rumor representations via the propagation structure. 7) \textbf{PLAN}: A transformer-based model \cite{khoo2020interpretable} for rumor detection to model long-distance interactions between any pair of tweets even irrelevant. 8) \textbf{HD-\textsc{Trans}}: \citet{ma2020debunking} proposed a model based on tree-transformer networks, which focuses on proving its effectiveness on shallow and deep conversations of datasets separately. Thus we compare it on TWITTER-S/-D. 9) \textbf{Bi-GCN}: A GCN-based model~\cite{bian2020rumor} on directed conversation trees to learn higher-level representations.

We use accuracy and class-specific F-measure as evaluation metrics. To make a fair comparison, we conduct five-fold cross-validation on the datasets following all baselines to obtain robust results.

{\setlength{\abovecaptionskip}{-0.1cm}
\setlength{\belowcaptionskip}{-0.1cm}
\begin{table*}[t]\small
\centering
\begin{center}
\resizebox{0.97\textwidth}{!}{
\begin{tabular}{l||c|cccc||ccccc}
\hline
Dataset                 & \multicolumn{5}{c|}{TWITTER-S}                                                                                                                                       & \multicolumn{5}{c}{TWITTER-D}                                                                                  \\ \hline
\multirow{2}{*}{Method} & \multirow{2}{*}{Acc.}               & NR                                 & FR                                 & TR             & UR                                  & \multicolumn{1}{c|}{\multirow{2}{*}{Acc.}} & NR             & FR             & TR             & UR             \\ \cline{3-6} \cline{8-11} 
                        &                                     & $\emph{F}_1$                                 & $\emph{F}_1$                                 & $\emph{F}_1$             & $\emph{F}_1$                                  & \multicolumn{1}{c|}{}                      & $\emph{F}_1$             & $\emph{F}_1$             & $\emph{F}_1$             & $\emph{F}_1$             \\ \hline \hline
DTR                     & 0.467                               & 0.622                              & 0.329                              & 0.520          & 0.299                               & \multicolumn{1}{c|}{0.566}                 & 0.447          & 0.577          & 0.555          & 0.484          \\
DTC                     & 0.523                               & 0.728                              & 0.418                              & 0.512          & 0.349                               & \multicolumn{1}{c|}{0.538}                 & 0.758          & 0.516          & 0.332          & 0.381          \\
RFC                     & 0.599                               & 0.782                              & 0.470                              & 0.561          & 0.385                               & \multicolumn{1}{c|}{0.582}                 & 0.774          & 0.501          & 0.461          & 0.395          \\ \hline
SVM-TK                  & 0.719                               & 0.705                              & 0.683                              & 0.785          & 0.682                               & \multicolumn{1}{c|}{0.669}                 & 0.698          & 0.649          & 0.689          & 0.615          \\
GRU-RNN                 & 0.715                               & 0.700                              & 0.697                              & 0.780          & 0.620                               & \multicolumn{1}{c|}{0.646}                 & 0.645          & 0.624          & 0.714          & 0.598          \\
RvNN                    & 0.749                               & 0.724                              & 0.729                              & 0.830          & 0.684                               & \multicolumn{1}{c|}{0.705}                 & 0.725          & 0.677          & 0.759          & 0.656          \\
PLAN                    & 0.764                               & 0.742                              & 0.744                              & 0.840          & 0.699                               & \multicolumn{1}{c|}{0.719}                 & 0.746          & 0.708          & 0.760          & 0.646          \\
HD-\textsc{Trans}                & 0.789                               & 0.749                              & \multicolumn{1}{l}{0.784}          & 0.837          & 0.776                               & \multicolumn{1}{c|}{0.768}                 & 0.773          & 0.781          & 0.783          & 0.721          \\
Bi-GCN                  & 0.790                               & 0.716                              & 0.758                              & 0.843          & 0.816                               & \multicolumn{1}{c|}{0.803}                 & 0.792          & 0.788          & 0.796          & 0.814          \\
\hline
ClaHi-GAT               & \multicolumn{1}{l|}{\textbf{0.847}} & \multicolumn{1}{l}{\textbf{0.806}} & \multicolumn{1}{l}{\textbf{0.817}} & \textbf{0.886} & \multicolumn{1}{l|}{\textbf{0.854}} & \multicolumn{1}{c|}{\textbf{0.835}}                             & \textbf{0.832} & \textbf{0.823} & \textbf{0.824} & \textbf{0.849} \\ \hline
\end{tabular}}
\end{center}
\caption{Rumor detection results on TWITTER datasets.}
\vspace{-0.2cm}
\label{tab:twitter}
\end{table*}}

\subsection{Rumor Classification Performance}

Table \ref{tab:twitter} and Table \ref{tab:pheme} show the performance of our proposed method 
versus all the compared methods on the TWITTER and PHEME datasets, where the best result of each column is bolded to indicate the significant improvement over all baselines ($p < 0.05$). To fairly compare with HD-\textsc{Trans}, our main experiments are conducted on TWITTER-S/-D and we also provide experimental results on the original TWITTER datasets in the appendix for completeness.

It is observed that the performances of the baselines in the first group based on handcrafted features are obviously poor. RFC performs relatively better because of the usage of additional temporal traits. Except for the first group, other baselines exploit the collective wisdom of the community by applying natural language processing to comments directed toward a claim without dependency on metadata and laborious feature engineering. 

Among the baselines without feature engineering in the second group, due to the representation power of message-passing architectures and tree structures, PLAN, HD-\textsc{Trans} and Bi-GCN outperform RvNN in general. However, our aggregation-based method achieves superior performance among all the baselines on different datasets, even in the case where data is just shallow/deep conversation separately or unbalanced, which reflects its keen judgment on rumors and indicates the flexibility of our model on different types of datasets. Different from the aforementioned baselines, ClaHi-GAT is based on the interaction topology considering not only the intrinsic structural property but also the interaction between close associated posts.

The outstanding results indicate that the claim-guided hierarchical attention mechanism based on undirected interaction graphs modeling can effectively enhance the representation learning using semantic and structural information.

\subsection{Ablation Study}
We perform ablation studies by discarding some important components of ClaHi-GAT on Twitter15\&16, and PHEME respectively, which include 1) \textit{ClaHi-GAT/DT}: Instead of the undirected interaction graph, we use the directed trees~\cite{ma2018rumor, bian2020rumor} as the model input. 2) \textit{GAT+EA+SC}: We simply concatenate the features of the claim with the node features at each GAT layer, to replace the claim-aware representation in Eq.\ref{equ7}. 3) \textit{w/o EA}: We discard the event-level (inference-based) attention as presented in Eq.\ref{equ10}. 4) \textit{w/o PA}: We neglect the post-level (claim-aware) attention by leaving out the gating module introduced in Eq.\ref{equ7}. 5) \textit{GAT}: The backbone model described in Sec.\ref{GATs}. 6) \textit{GCN}: The vanilla graph convolutional networks with no attention. 

{\setlength{\abovecaptionskip}{-0.1cm}
\setlength{\belowcaptionskip}{-0.1cm}
\begin{table}[t]\small
\centering
\begin{center}
\resizebox{0.45\textwidth}{!}{
\begin{tabular}{l|c|cc}
\hline
\multirow{2}{*}{Method} & \multirow{2}{*}{Acc.} & Non-rumor      & Rumor          \\ \cline{3-4} 
                        &                       & $\emph{F}_1$             & $\emph{F}_1$             \\ \hline \hline
DTR                     & 0.657                 & 0.772          & 0.317          \\
DTC                     & 0.670                 & 0.755          & 0.494          \\
RFC                     & 0.709                 & 0.809          & 0.393          \\ \hline
SVM-TK                  & 0.785                 & 0.839          & 0.677          \\
GRU-RNN                 & 0.775                 & 0.832          & 0.658          \\
RvNN                    & 0.829                 & 0.873          & 0.736          \\
PLAN                    & 0.824                 & 0.868          & 0.731          \\
Bi-GCN                  & 0.835                 & 0.872          & 0.764          \\\hline
ClaHi-GAT               & \textbf{0.859}        & \textbf{0.893} & \textbf{0.790} \\ \hline
\end{tabular}}
\end{center}
\caption{Rumor detection results on PHEME dataset.}
\vspace{-0.3cm}
\label{tab:pheme}
\end{table}}

As demonstrated in Table \ref{ablation}, \textit{ClaHi-GAT/DT} suffers a large decrease, indicating that our proposed undirected interaction graph modeling contributes to the final performance and its combination with claim-guided hierarchical graph attention encoding is critical. Each component of our model alone improves the model, indicating their effectiveness for embedding the interaction graph. Specifically, \textit{GAT} makes remarkable improvements over \textit{GCN}, reflecting the role of naive attention in reducing the weights of noisy nodes; \textit{w/o EA} and \textit{w/o PA} consistently outperform \textit{GAT}, suggesting that both levels of attention are comparably helpful; Combining them hierarchically makes further improvements and implies their complementary as represented by \textit{ClaHi-GAT}, and replacing the claim-aware attention at the post level with simple concatenation (\textit{GAT+EA+SC}) also leads to performance degradation, reaffirming the more effective and reasonable involvement of claims and advantages of the claim-guided hierarchical attention mechanism.

\subsection{Evaluation of Undirected Interaction Graphs}

We present more qualitative analyses about the undirected interaction graph in this section. Figure~\ref{fig:ablation_modeling} provides the experimental results of ClaHi-GAT and the following models based on different modeling ways:
\begin{enumerate}
\item \textbf{ClaHi-GAT/DT} Utilize the directional tree applied in past influential works~\cite{ma2018rumor,ma2020debunking,bian2020rumor} as the modeling way instead of our proposed undirected interaction graph.

{
\setlength{\belowcaptionskip}{-0.1cm}
\begin{table}[t]
\centering
\resizebox{0.45\textwidth}{!}{
\begin{tabular}{l|ccc}
\hline
\multirow{2}{*}{Method} & Twitter15 & Twitter16 & PHEME \\ \cline{2-4} 
                        & Acc.      & Acc.      & Acc.  \\ \hline \hline
ClaHi-GAT               & 0.891     & 0.908     & 0.859 \\ \hline
ClaHi-GAT/DT            & 0.813     & 0.848     & 0.837 \\
GAT+EA+SC               & 0.853     & 0.866     & 0.846 \\
GAT+PA(w/o EA)          & 0.878     & 0.889     & 0.848 \\
GAT+EA(w/o PA)          & 0.847     & 0.864     & 0.845  \\ 
GAT                     & 0.835     & 0.854     & 0.840 \\
GCN                     & 0.825     & 0.820     & 0.832 \\ \hline
\end{tabular}}
\caption{Ablation studies on our proposed model.}
\vspace{-0.3cm}
\label{ablation}
\end{table}}

\item \textbf{ClaHi-GAT/DTS} Based on the directional tree structure similar to ClaHi-GAT/DT but the explicit interactions between sibling nodes are taken into account.

\item \textbf{ClaHi-GAT/UD} The modeling way is our undirected interaction topology but without considering the explicit correlations between sibling nodes that reply to the same target.

\item \textbf{ClaHi-GAT} In this paper, we propose to model the conversation thread as an undirected interaction graph for our claim-guided hierarchical graph attention networks. 
\end{enumerate}

From the experimental results of Figure~\ref{fig:ablation_modeling}, we draw the following observations:

\textbf{Effectiveness of exploring coherent opinions among sibling nodes.} Compared with \textit{ClaHi-GAT/DT}, \textit{ClaHi-GAT/DTS} achieves $0.8\%$, $0.6\%$ and $0.5\%$ boosts in accuracy on Twitter15, Twitter16 and PHEME respectively. Compared with \textit{ClaHi-GAT/UD}, \textit{ClaHi-GAT} achieves $5.6\%$, $4.3\%$ and $1.1\%$ boosts in accuracy on Twitter15, Twitter16 and PHEME respectively. It proves the effectiveness of the enhanced interaction of user opinions by exploring the correlation between sibling nodes that reply to the same target. 

{
\setlength{\belowcaptionskip}{-0.2cm}
\begin{figure}[t]
    \centering
    \setlength{\belowcaptionskip}{0.1cm}
    \resizebox{0.49\textwidth}{!}{\includegraphics{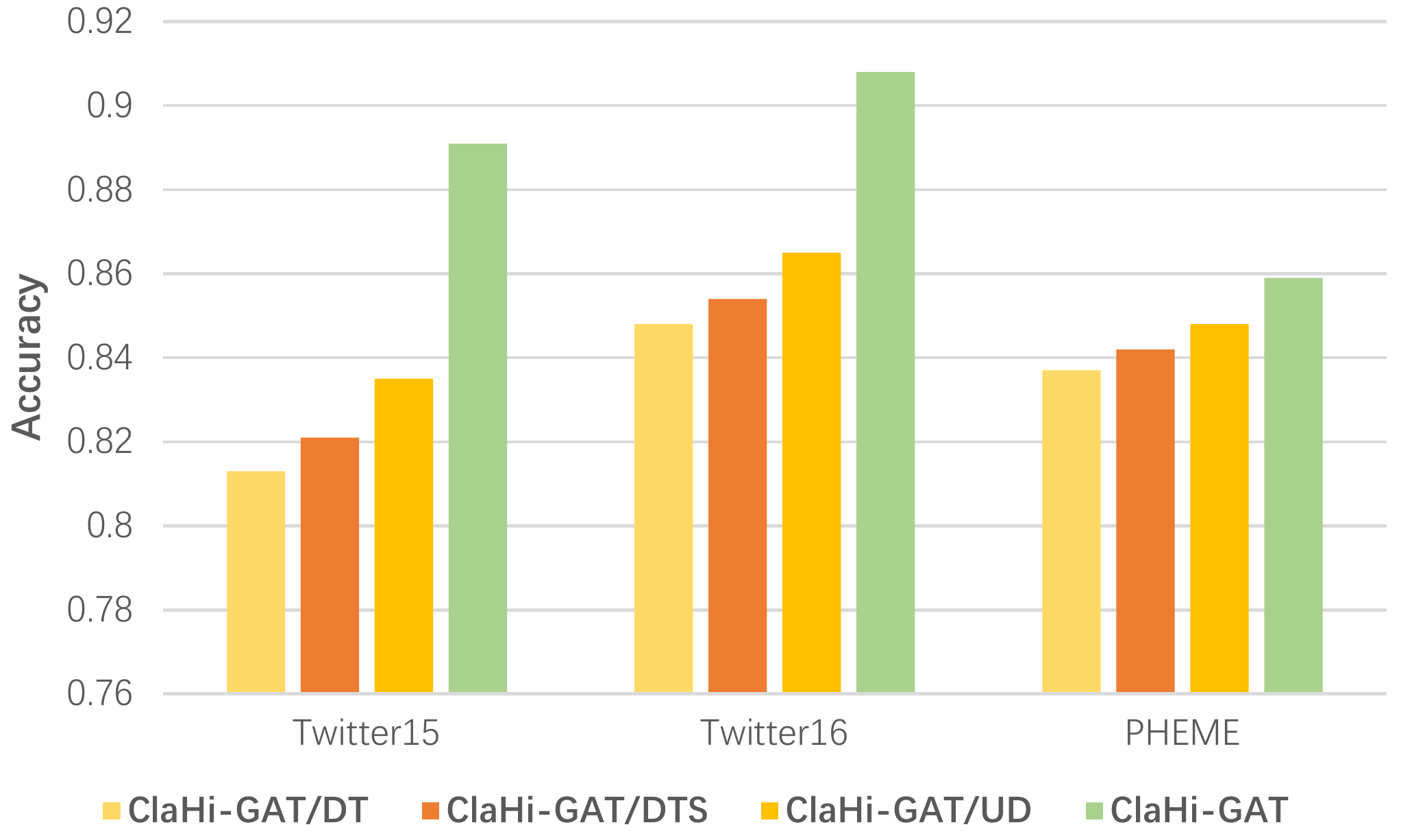}}
    \caption{The rumor classification performance of ClaHi-GAT based on different modeling ways.}
    \label{fig:ablation_modeling}
\vspace{-0.7cm}
\end{figure}}

\textbf{Effectiveness of the undirected graphs.} Due to the simplex interactions between posts in the directional tree, the interaction between sibling nodes can not have a strong impact. Therefore, we propose the undirected structure to strengthen the aggregation of rumor indication features and maximize the influence of the interaction between sibling nodes. We can see that without considering the sibling relationship, \textit{ClaiHi-GAT/UD} has better results than \textit{ClaHi-GAT/DT}, suggesting that the combination of the undirected graph with our proposed claim-guided hierarchical graph attention mechanism is more suitable and complementary. Not only that, \textit{ClaHi-GAT} boosts the performance as compared with \textit{ClaHi-GAT/DTS}, showing $7.0\%$, $5.4\%$ and $1.7\%$ improvements in accuracy on the three datasets, which reveals that the undirected interaction topology does enhance semantic associations and fusion.

{\setlength{\abovecaptionskip}{-0.1cm}
\setlength{\belowcaptionskip}{-0.1cm}
\begin{figure*}[ht]
\centering
\subfigure[Twitter15 (posts count)]{
\begin{minipage}[t]{0.33\linewidth}
\centering
\scalebox{0.85}{\includegraphics[width=6cm]{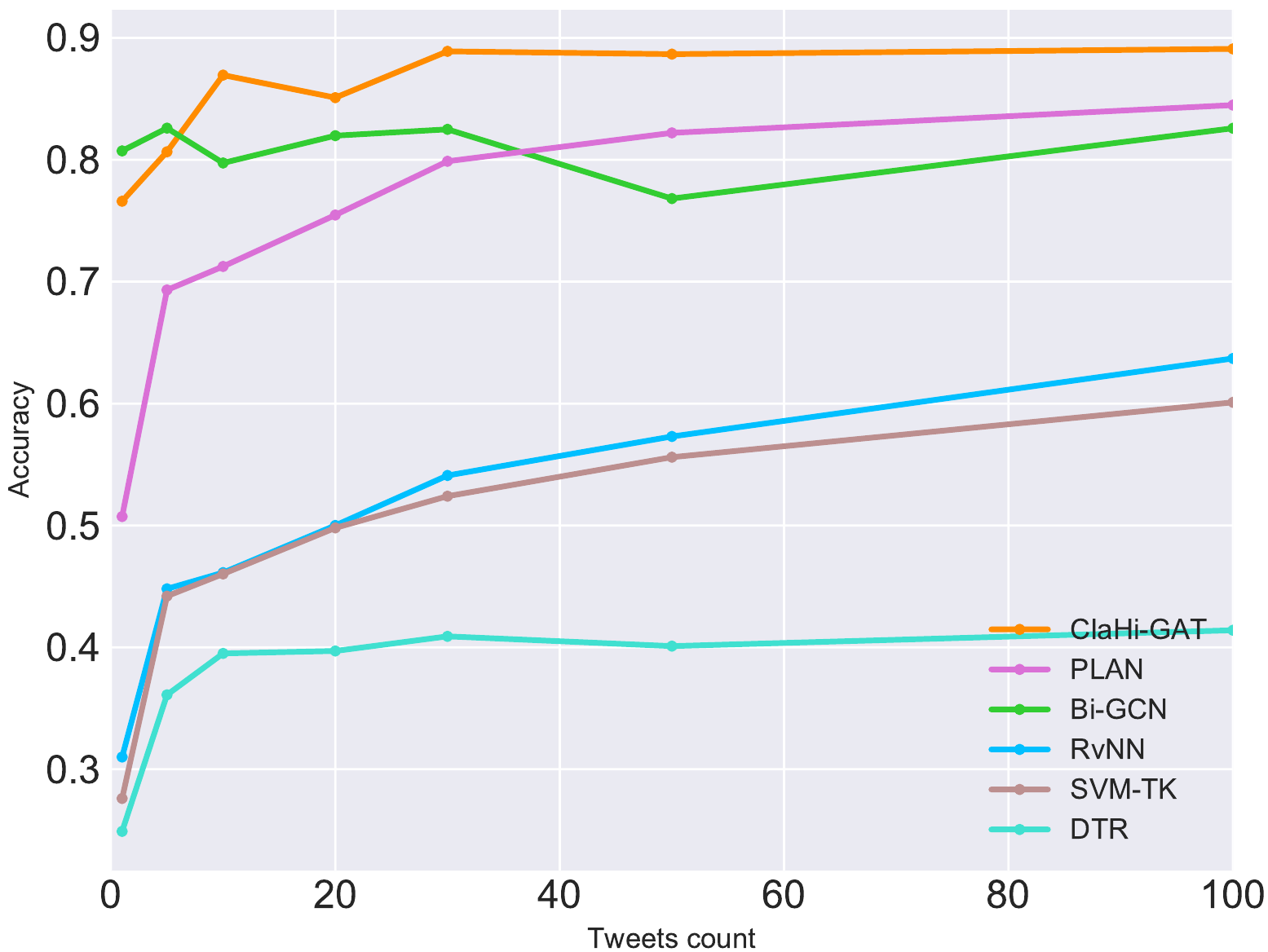}}
\label{fig:twitter15_early}
\end{minipage}%
}%
\subfigure[Twitter16 (posts count)]{
\begin{minipage}[t]{0.33\linewidth}
\centering
\scalebox{0.85}{\includegraphics[width=6cm]{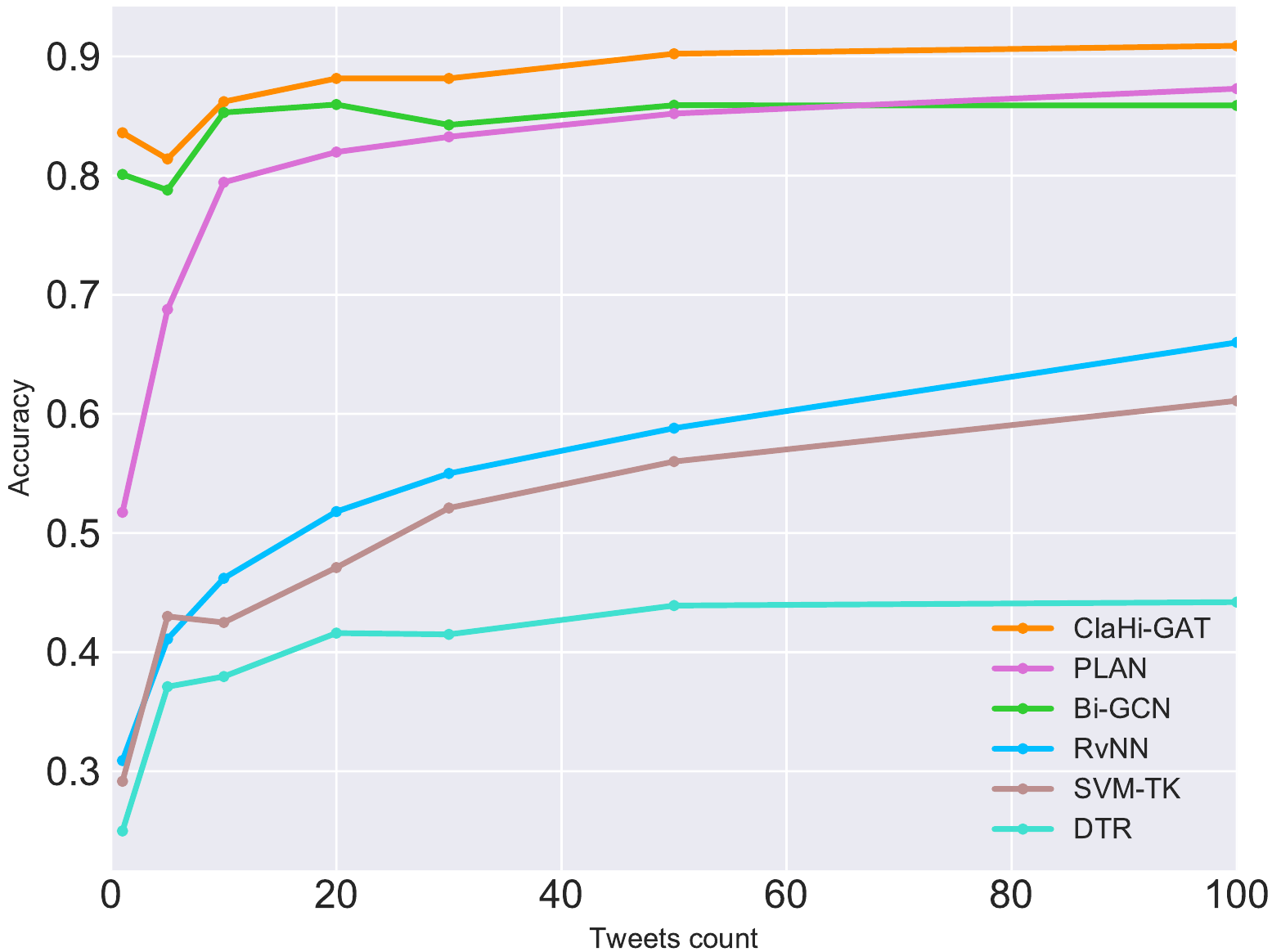}}
\label{fig:twitter16_early}
\end{minipage}%
}%
\subfigure[PHEME (elapsed time)]{
\begin{minipage}[t]{0.33\linewidth}
\centering
\scalebox{0.85}{\includegraphics[width=6cm]{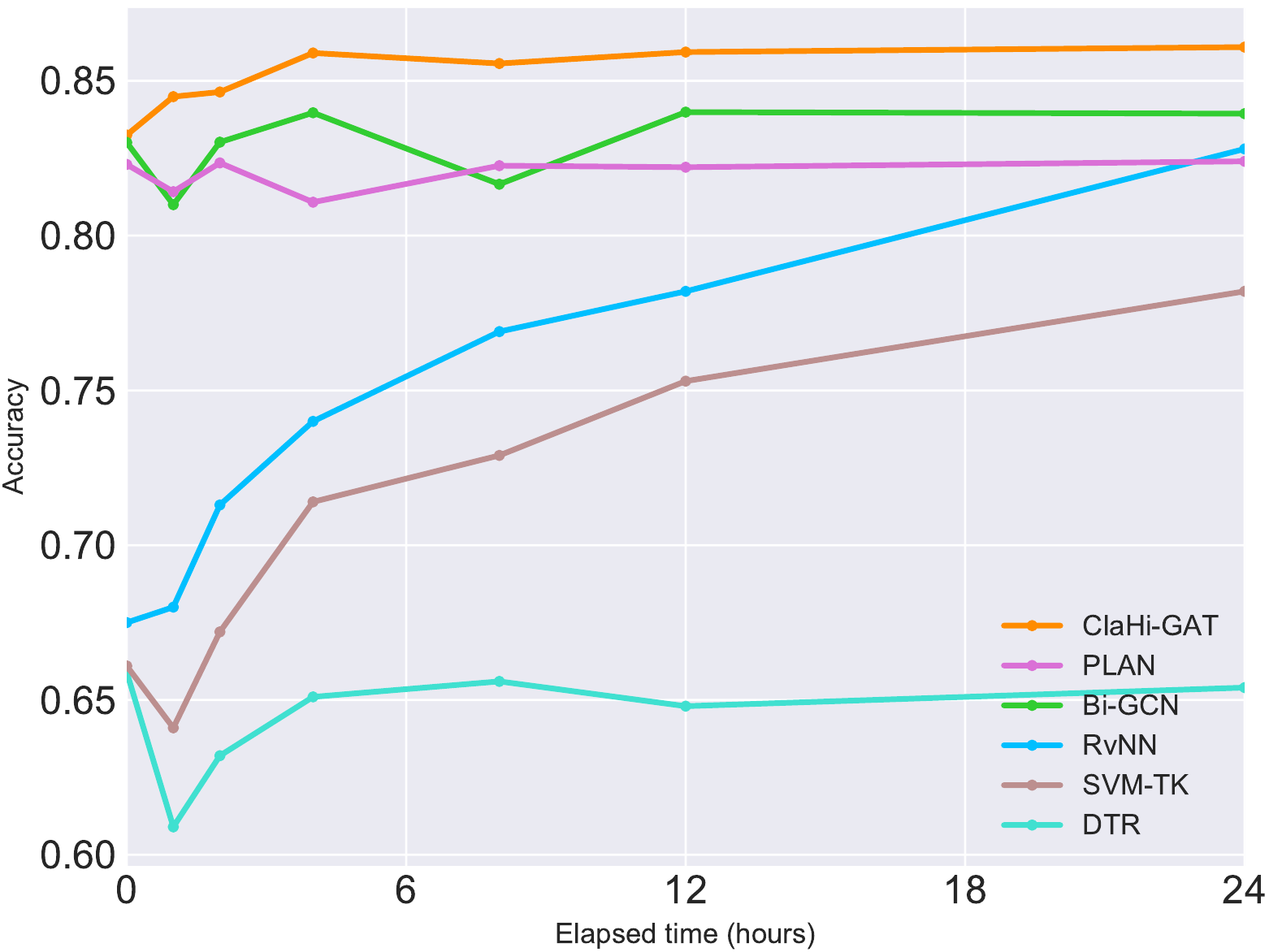}}
\label{fig:pheme_early}
\end{minipage}%
}%
\centering
\caption{Early rumor detection accuracy at different checkpoints in terms of post count (or elapsed time) on Twitter15, Twitter16, and PHEME datasets.}
\label{fig:early_detection}
\vspace{-0.2cm}
\end{figure*}}

\subsection{Early Rumor Detection}
To take preventive measures to rumor spreading in a timely manner, debunking rumors at the early stage of their propagation is important. In early detection task, we compare different detection methods at a series of checkpoints of ``delays" that can be measured by either the count of responsive posts received (for Twitter15\&16 dataset) or the time elapsed since the claim was posted (for PHEME dataset). The performance is evaluated by the accuracy obtained when we incrementally scan test data in order of time until the target time delay or post volume is reached. 

{
\setlength{\belowcaptionskip}{-0.2cm}
\begin{figure}[t]
    \centering
    \setlength{\belowcaptionskip}{0.1cm}
    \resizebox{0.49\textwidth}{!}{\includegraphics{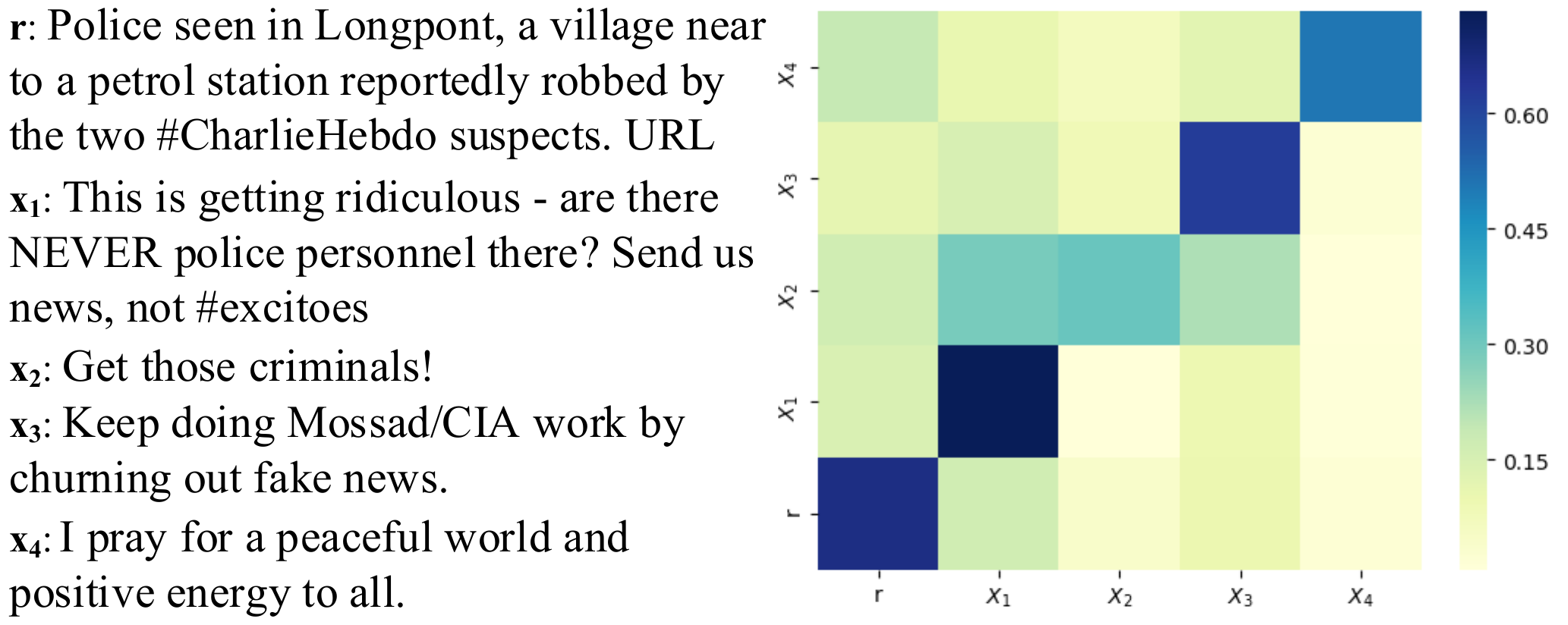}}
    \caption{Example of correctly detected false rumors at early stage of our model.}
    \label{fig:case}
\vspace{-0.7cm}
\end{figure}}

Figure~\ref{fig:early_detection} shows the performances of our ClaHi-GAT method versus PLAN, Bi-GCN, RvNN, SVM-TK, and DTR at various deadlines. It is observed that models leveraging the structural information (e.g., ClaHi-GAT method, PLAN, and Bi-GCN) reach relatively high accuracy at a very early period after the initial broadcast. One interesting phenomenon is that the early performance of all methods fluctuated more or less. We conjecture that this is because with the propagation of the claim there is more semantic and structural information, meanwhile, the noisy information is increased. Therefore, the results show that our model is insensitive to data and has better stability and robustness. ClaHi-GAT only needs about 30 posts on TWITTER and around 4 hours on PHEME, to achieve the saturated performance, which indicates remarkably superior early detection performance of our method. 

To get an intuitive understanding of what is happening when we use the ClaHi-GAT model, we present an example of sibling nodes responding to the false claim $r$ in our undirected interaction graph with a heatmap of the averaged multi-head attention score of neighbors at the last graph attention layer. In Figure~\ref{fig:case} we can see that for the false rumor, the inaccurate information like $x_2$ and $x_4$ could reduce their weights and pay more attention to the claim-related denial or questioning posts that contradict the claim, which may help us correctly predict the false rumor. Furthermore, the obtained attention scores play a crucial role in the interpretability of the prediction by the highlighted informative posts and hidden correlations.

\section{Conclusion}
In this paper, we propose a novel Claim-guided Hierarchical Graph Attention Network based on undirected interaction graphs to learn graph attention-based embeddings that attend to user interactions for rumor detection. 
Multi-level rumor indicative features could be better captured via the claim-aware attention at post level and the inference-based attention at event level. 
The results on three public benchmark datasets confirm the advantages of our model. Our framework is expected to provide new guidance for future rumor detection work.

\section*{Acknowledgements}
We thank all anonymous reviewers for their helpful comments and suggestions. This work was partially supported by the Foundation of Guizhou Provincial Key Laboratory of Public Big Data (No.2019BDKFJJ002). Jing Ma was supported by HKBU direct grant (Ref. AIS 21-22/02).

\bibliography{anthology,custom}
\bibliographystyle{acl_natbib}
\newpage
\appendix

\section{Dataset Details}
We conduct experiments on three public benchmark datasets, including Twitter15 \cite{ma2017detect}, Twitter16 \cite{ma2017detect}, and PHEME \cite{zubiaga2016learning}. Twitter15 and Twitter16 datasets contain four labels: Non-rumor (NR), False Rumor (FR), True Rumor (TR), and Unverified Rumor (UR), while the PHEME dataset contains two binary labels: Rumor and Non-rumor. The statistics of the three datasets are shown in Table \ref{tab:statistics}.

\section{Implementation Details}
During model training, we exploit the cross-entropy loss of the predictions and ground truth distributions over training data with the L2-norm. We set the number L of the graph attention layer as 2, and the head number K as 4. Parameters are updated through back-propagation \cite{collobert2011natural} with the Adam optimizer \cite{kingma2014adam}. The learning rate is initialized as 0.0005, and the dropout rate is 0.2. Early stopping \cite{yao2007early} is applied to avoid overfitting. We run all of our experiments on one single NVIDIA Tesla V100-PCIE GPU. We set the batch size to 128. Since the focus in this paper is primarily on better leveraging the graph structure and correlations between nodes, we choose the text representations widely used in previous works~\cite{ma2020debunking, ma2020attention}. Specifically, we use the GLOVE 300d \cite{pennington2014glove} embedding to represent each token in a tweet and get 128-dimensional contextual sentence features with a single-layer Bi-LSTM encoder. The hidden dimension of each node is set to 128. We hold out $10\%$ of the datasets for tuning the hyperparameters and conduct 5-fold cross-validation on the rest of the datasets. We use accuracy and class-specific F-measure as evaluation metrics. The average runtime for our approach on five-fold cross-validation in one iteration is about 1.0 hours. The number of total parameters is 52,851,029 for our model. We implement our model with pytorch\footnote{\url{pytorch.org}}.

\begin{table}[t]
\centering
\begin{center}
\resizebox{0.45\textwidth}{!}{
\begin{tabular}{l|ccc}
\hline
Statistic               & Twitter15 & Twitter16 & PHEME    \\ \hline \hline
\# of source tweets     & 1,490     & 818       & 5,802    \\ \hline
\# of tree nodes        & 76,351    & 40,867    & 30,376   \\ \hline
\# of non-rumors        & 374       & 205       & 3,830    \\ \hline
\# of false rumors      & 370       & 205       & 1,972    \\ \hline
\# of true rumors       & 372       & 205       & –        \\ \hline
\# of unverified rumors & 374       & 203       & –        \\ \hline
Avg. time length / tree & 444 Hours & 196 Hours & 18 Hours \\ \hline
Avg. \# of posts / tree & 52        & 50        & 6        \\ \hline
\end{tabular}}
\end{center}
\caption{Statistics of TWITTER and PHEME Dataset.}
\label{tab:statistics}
\end{table}

\begin{table*}[t]
\centering
\begin{center}
\resizebox{0.9\textwidth}{!}{
\begin{tabular}{l|c|cccc|ccccc}
\hline
Dataset                 & \multicolumn{5}{c|}{Twitter15}                                                            & \multicolumn{5}{c}{Twitter16}                                                                                  \\ \hline
\multirow{2}{*}{Method} & \multirow{2}{*}{Acc.} & NR             & FR             & TR             & UR             & \multicolumn{1}{c|}{\multirow{2}{*}{Acc.}} & NR             & FR             & TR             & UR             \\ \cline{3-6} \cline{8-11} 
                        &                       & $\emph{F}_1$             & $\emph{F}_1$             & $\emph{F}_1$             & $\emph{F}_1$             & \multicolumn{1}{c|}{}                      & $\emph{F}_1$             & $\emph{F}_1$             & $\emph{F}_1$             & $\emph{F}_1$             \\ \hline \hline
DTR                     & 0.409                 & 0.501          & 0.311          & 0.364          & 0.473          & \multicolumn{1}{c|}{0.414}                 & 0.394          & 0.273          & 0.630          & 0.344          \\
DTC                     & 0.454                 & 0.415          & 0.355          & 0.733          & 0.317          & \multicolumn{1}{c|}{0.465}                 & 0.643          & 0.393          & 0.419          & 0.403          \\
RFC                     & 0.565                 & 0.810          & 0.422          & 0.401          & 0.543          & \multicolumn{1}{c|}{0.585}                 & 0.752          & 0.415          & 0.547          & 0.563          \\ \hline
SVM-TK                  & 0.667                 & 0.619          & 0.669          & 0.772          & 0.645          & \multicolumn{1}{c|}{0.662}                 & 0.643          & 0.623          & 0.783          & 0.655          \\
GRU-RNN                 & 0.641                 & 0.684          & 0.634          & 0.688          & 0.571          & \multicolumn{1}{c|}{0.633}                 & 0.617          & 0.715          & 0.577          & 0.527          \\
RvNN                    & 0.723                 & 0.682          & 0.758          & 0.821          & 0.654          & \multicolumn{1}{c|}{0.737}                 & 0.662          & 0.743          & 0.835          & 0.708          \\
Bi-GCN                  & 0.826                 & 0.779          & 0.835          & 0.888          & 0.791          & \multicolumn{1}{c|}{0.859}                 & 0.773          & 0.857          & 0.930          & 0.860          \\
PLAN                    & 0.845                 & 0.823          & 0.858          & 0.895          & 0.802          & \multicolumn{1}{c|}{0.874}                 & 0.853          & 0.839          & 0.917          & 0.888          \\ \hline
ClaHi-GAT               & \textbf{0.891}        & \textbf{0.878} & \textbf{0.882} & \textbf{0.931} & \textbf{0.867} & \multicolumn{1}{c|}{\textbf{0.908}}        & \textbf{0.862} & \textbf{0.916} & \textbf{0.954} & \textbf{0.901} \\ \hline
\end{tabular}}
\end{center}
\caption{Rumor detection results on original Twitter15 and Twitter16 datasets.}
\vspace{-0.2cm}
\label{tab:oritwitter}
\end{table*}

{
\setlength{\belowcaptionskip}{-0.2cm}
\begin{figure*}[t]
    \centering
    \setlength{\belowcaptionskip}{0.1cm}
    \resizebox{0.73\textwidth}{!}{\includegraphics{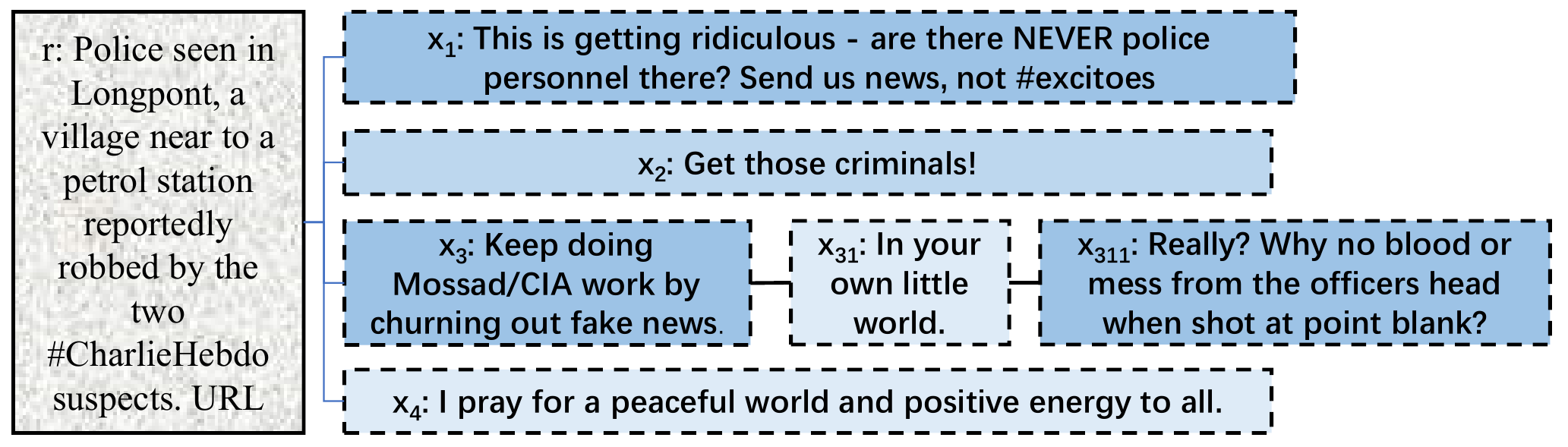}}
    \caption{A sample case of correctly detected false rumors of our model. We show important tweets in the conversation and truncate others.}
    \label{fig:case_entail}
\vspace{-0.7cm}
\end{figure*}}

\section{Supplemental Experiments}
We provide a supplemental experiment on the original version of TWITTER datasets for completeness, as depicted in Table~\ref{tab:oritwitter}. Previous works like \citet{yuan2019jointly}, \citet{lu2020gcan} and \citet{huang2020heterogeneous} also conducted on the original TWITTER datasets leveraging the bias and social network of the source of the claim. We did not include these models in our experiments, because: 1) In this paper, we work on detecting rumors solely from the posts and comments, which takes advantage of the ``wisdom of crowds" information by mining conflicting viewpoints in microblogs. In order to improve the performance of our model effectively and equitably, we do not leverage the identities of user accounts or characteristics. 2) The experimental setups for the three models are not consistent with 5-fold cross-validation and even use the pre-split train, valid and test datasets by themselves, which can not easily conduct a fair comparison with the performance on 5-fold cross-validation for all baselines and our proposed model. Here we also do not include \textsc{HD-Trans} in our supplemental experiments, because it focuses on proving its effectiveness on the shallow and deep trees separately instead of the original TWITTER datasets. Our implementation of the code\footnote{\url{https://github.com/TianBian95/BiGCN}} released by Bi-GCN has a big gap compared with results reported in their paper~\cite{bian2020rumor}, though our model still performs better due to the robustness in five-fold cross-validation. The results indicate that our proposed methods outperform all the baselines, confirming the advantages of ClaHi-GAT for rumor detection task.

\section{Case Study}
For a more comprehensive analysis on the event-level attention, we present an example of correctly detected false rumors, whose nodes are colored with the inference-based attention scores (i.e., `$\beta_i$' in Eq.10 of the main body of this paper) at the event level (the higher the score, the darker the color). The visualization of tweets in Figure~\ref{fig:case_entail} shows that the ClaHi-GAT captures informative tweets in the conversation, which have a contradiction relation towards the false claim. Hence, our event-level attention module can notice salient indicators of rumor veracity in the conversation thread, e.g., posts that contradict the false claims or entail the true claims, and then combine them to give a correct prediction.

\section{Future Work}
We will explore the following directions in the future based on error cases where our model can not predict the correct label of the claim:
\begin{enumerate}
\item Traditional embedding methods like static word vectors (e.g., GloVe or Word2Vec) used in this paper cannot disambiguate homonyms, express semantic and syntactic patterns well, especially casual expression in writing on social media. Representation from Transformer pre-training may effectively help us learn more context-aware representation at the token level. We will explore how to inject the generalized contextual information via pre-trained language models into our proposed framework, to further investigate the performance improvement.

\item The event-level attention component attempts to investigate the inference relationship between a claim and its responsive post. One issue of such component is the lack of explicit supervision signal of recognizing textual inference patterns. In the future, we will utilize some existing language inference datasets with explicit labels to obtain some prior knowledge to tackle this challenge. Specifically, the knowledge of recognizing entailment relations in the trained model can be transferred to our target component.

\item In reality, some users tend to simply reshare a claim without expressing their opinions or comments. Our model cannot perfectly handle the instance that few users’ engagements are available. That case is similar to the early rumor detection scenario. Although our model achieves superior performance on the early rumor detection task, it still suffers from incorrect prediction caused by the situation where users just mainly retweet the claim without more opinion expression. Also, we found an attractive point is that the same user might reply to their own claim in the propagation way. It would be heuristic for us to model novel social networks considering the special modes (e.g., retweet or reply by the node itself posting the claim) during the rumor propagation.
\end{enumerate}

\end{document}